\documentclass[10pt,twocolumn,letterpaper]{article}

\usepackage{iccv}
\usepackage{times}
\usepackage{epsfig}
\usepackage{graphicx}
\usepackage{amsmath}
\usepackage{amssymb}
\usepackage{adjustbox}

\usepackage[breaklinks=true,bookmarks=false]{hyperref}

\iccvfinalcopy 

\begin{document}

\title{CLRmatchNet: Enhancing Curved Lane Detection with Deep Matching Process}

\author{Sapir Kontente\\
{\tt\small sapirk@mail.tau.ac.il}
\and
Roy Orfaig\\
{\tt\small royorfaig@tauex.tau.ac.il}
\and 
Ben-Zion Bobrovsky \\
{\tt\small bobrov@tauex.tau.ac.il}\\
\and
{School of Electrical Engineering, Tel-Aviv University}
}

\maketitle

\begin{abstract}
Lane detection plays a crucial role in autonomous driving by providing vital data to ensure safe navigation. Modern algorithms rely on anchor-based detectors, which are then followed by a label-assignment process to categorize training detections as positive or negative instances based on learned geometric attributes. Accurate label assignment has great impact on the model performance, that is usually relying on a pre-defined classical cost function evaluating GT-prediction alignment. However, classical label assignment methods face limitations due to their reliance on predefined cost functions derived from low-dimensional models, potentially impacting their optimality. Our research introduces MatchNet, a deep learning submodule-based approach aimed at improving the label assignment process. Integrated into a state-of-the-art lane detection network such as the Cross Layer Refinement Network for Lane Detection (CLRNet), MatchNet replaces the conventional label assignment process with a submodule network. The integrated model, CLRmatchNet, surpasses CLRNet, showing substantial improvements in scenarios involving curved lanes, with remarkable improvement across all backbones of +2.8\% for ResNet34, +2.3\% for ResNet101, and +2.96\% for DLA34. In addition, it maintains or even improves comparable results in other sections. Our method boosts the confidence level in lane detection, allowing an increase in the confidence threshold. Our code is available at: \href{https://github.com/sapirkontente/CLRmatchNet.git}{https://github.com/sapirkontente/CLRmatchNet.git}
\end{abstract}

\begin{figure}[t]
    \begin{center}
    \includegraphics[width=1\linewidth]{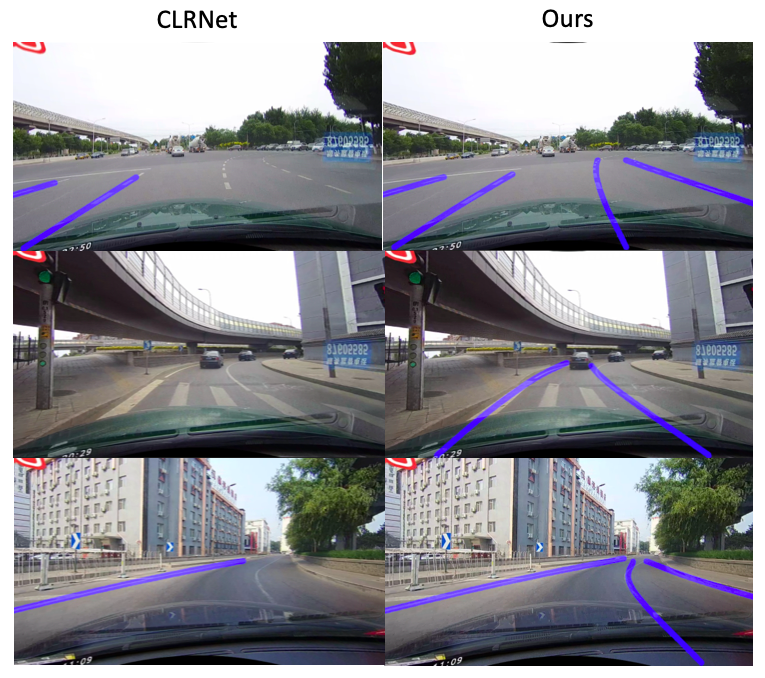}
    \end{center}
    \caption{Visualizing lane detection results: a comparison between CLRNet and CLRmatchNet (our approach) of CULane testing set in the curve category}
    \label{fig:bestresult}
\end{figure}

\section{Introduction}
\label{sec:intro}
Label assignment is a critical element within modern object detection models that has a substantial impact on their overall performance. The methodologies used for label assignment can yield significantly diverse results in the context of detection models. During this phase, the main objective is to establish a meaningful association between each prediction and its corresponding ground-truth (GT) for classification purposes. When a prediction aligns with a GT, it is considered as positive, whereas predictions lacking relevance to any GT are labeled as negatives. During training, the model learns to assign high confidence scores to the lanes labeled as positive; therefore, predicted lanes that have a high alignment score with the GTs are expected to receive high classification scores at the test time. Label assignment serves as a stabilizing factor in the training process. Instead of optimizing all predictions, label assignment allows us to filter the most promising predictions for subsequent fine-tuning. This strategy allows the model to focus on the strong predictions without being affected by the weaker ones.

Typically, this procedure incorporates a predefined cost function that takes into account various geometric attributes and classification scores, assigning a score to each GT-prediction pairing. A filtering mechanism is then applied to select the most relevant pairings according to their respective scores. Modern object detectors employ a variety of cost functions, some relying solely on geometric metrics while others incorporate classification scores. Earlier models used fixed thresholds based on cost function scores to distinguish between positive and negative predictions, whereas recent models employ adaptive label assignment strategies that dynamically calculate thresholds for improved performance. 

The formulation of the cost function is of utmost importance, as it directly determines the diagnostic ability of the distribution of matching degrees between pairs. A cost function based on classification scores and geometric attributes, such as the lane shape, lane curvature, start point, length, angle related to the bottom of the frames, and more, should be properly designed to avoid negatively affecting its diagnostic ability. Consequently, it is crucial to formulate the cost function in such a way that each of these attributes is effectively considered, enabling the label assignment process to identify the most relevant lanes and label them as positive. Our findings reveal that the current cost function utilized in recent lane detection models lacks optimality when it comes to curves. Through refinement of the model using a better-suited label assignment method, we demonstrate the potential for substantial improvement in the detection of curved lanes.

We introduce a novel approach for the label assignment process named MatchNet, a tiny neural network with the ability to learn a more accurate label assignment process while dynamically determining the number of predictions assigned to each GT. MatchNet has been incorporated into the state-of-the-art lane detection network, Cross Layer Refinement Network for Lane Detection (CLRNet) \cite{CLRNet}. In this integration, MatchNet replaces the conventional classic label assignment process with a submodule deep neural network. We refer to the full integrated model as CLRmatchNet. Fig. \ref{fig:bestresult} illustrates the results of our improved approach compared to CLRNet. Clearly, our method has a significant influence on the final lane detection, outperforming the initial CLRNet's results. Our main contributions can be summarized as follows:
\begin{itemize}
    \item \textit{\textbf{Deep learning-based label assignment}}: 
    Our work introduces an approach that aims to enhance models' performance by utilizing a submodule of a neural network mechanism for label assignment during training, instead of relying on classical cost functions that are low-dimensional and limited in challenging scenarios.
    \item \textit{\textbf{Enhanced performance of curved lanes}}: Our novel matching network significantly improves the detection capability of challenging lane types, such as curved lanes, while maintaining comparable performance across other testing categories.
    \item \textit{\textbf{Enhance confidence values of the true positive lane}}: Using MatchNet has substantially increased our confidence in lane detection, improving our ability to distinguish between true-positive and false-positive detections. First, low-confidence true predictions are now detected with greater confidence. Moreover, this enables us to set a higher final confidence threshold for detections, consequently minimizing false positive detections.
    \item \textit{\textbf{Optimized number of predictions selection}}: In the training phase, the label assignment module is responsible for matching the predictions with GTs in each image. It aims to selects the k most suitable matches for each GT, closely aligning with it, with a maximum of four matches allowed. Our approach optimizes the selection of k for each scenario utilizing MatchNet scores for each prediction-GT pair. Matches exceeding a predefined confidence threshold, are considered positives. 
\end{itemize}

\section{Related Work}
\label{sec:relatedwork}
Lane detection is a pivotal research area within the field of computer vision, playing a critical role in diverse applications such as autonomous driving and advanced driver-assistance systems. Numerous studies, including the recent researches \cite{ko2021key,zheng2022novel, qiu2022mfialane, li2023pga} underscore the significance of advancing lane detection methodologies to enhance the safety and efficiency of intelligent transportation systems. 
Lane detection faces challenges due to various lane configurations, driving scenarios, and environmental factors. Conventional object detection methods are not ideal because lanes are thin and elongated. Consequently, researchers have developed specific approaches to the task of lane detection. This section provides a summary of recent deep learning-based methods, categorized into segmentation-based, keypoint-based methods, parametric prediction methods and row-based methods according to their line shape description strategies.

\textbf{Segmentation-based Methods.} Segmentation-based methods approach lane detection as a pixel-wise classification task and require post processing operations to infer lane information. Early approaches such as SCNN \cite{CULane_F1} used a multi-class classification strategy, but this proved inflexible. To enhance instance accuracy, post-clustering strategies were widely adopted. Recent studies \cite{chougule2018reliable, qin2020ultra} highlight the inefficiency of describing lane lines as masks. In response, anchor-based methods and row-wise detection methods have been proposed to address these limitations.

\textbf{Keypoint-based Methods.} Keypoints methods detect critical points along lane boundaries, offering flexibility and adaptability to complex road conditions by inferring lane shapes from these points and grouping them into complete lanes through post-processing. PINet \cite{ko2021key} employing key point estimation and instance segmentation, with multiple stacked hourglass networks trained simultaneously, allowing for flexible model sizing based on the computing power of the target environment. FOLOLane \cite{qu2021focus} focuses on modeling local patterns and predicting global structures in a bottom-up approach using separate heads of a CNN to detect keypoints and refine their locations. GANet \cite{wang2022keypoint} directly regresses the key-points to the starting points of lane lines, enabling parallel processing for efficient association, while incorporating a Lane-aware Feature Aggregator (LFA) to capture local correlations and enhance global associations. These approaches require a post-processing step to organize the lane points into lane instances, which incurs high computational costs.

\textbf{Parametric Prediction Methods.} Parametric prediction methods directly output parametric lines expressed by curve equations. PolyLaneNet \cite{polylanenet} was the first to propose a polynomial representation of each lane using deep polynomial regression. LSTR \cite{lstr} proposed an end-to-end method that directly outputs parameters of a lane shape model, using a network built with a transformer to learn richer structures and context. BSNet \cite{BSNet} exploit the b-spline curve to fit lane lines.
Parametric prediction methods have yet to surpass other approaches.

\begin{figure*}[t]
    \begin{center}
    \includegraphics[width=1\linewidth]{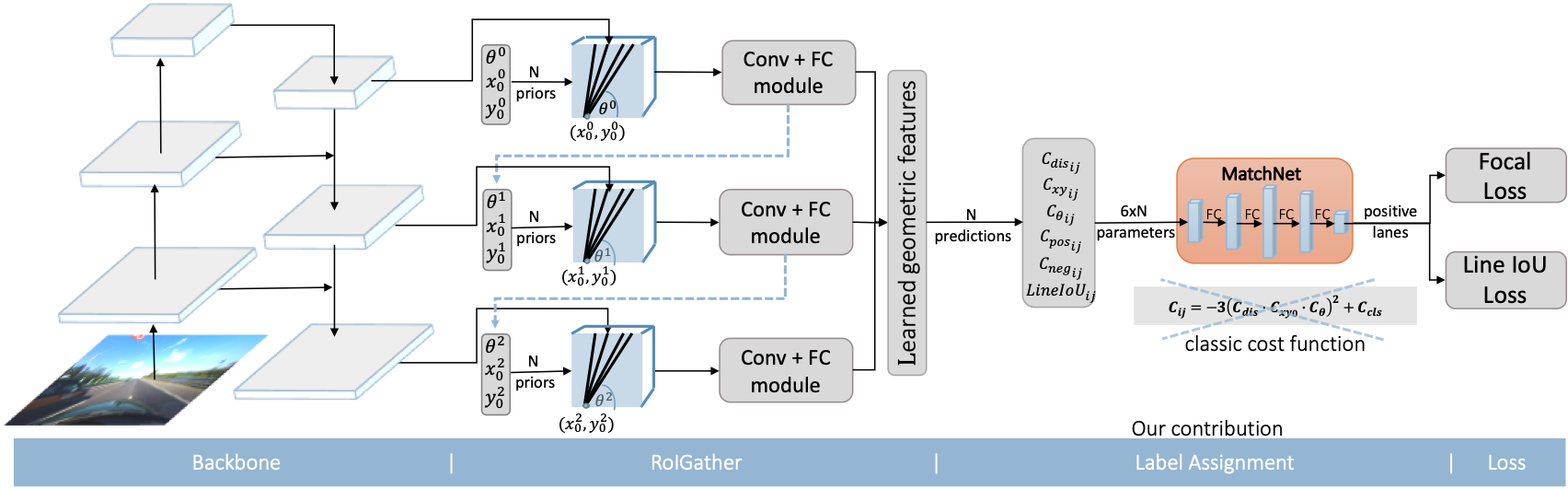}
    \end{center}
    \caption{CLRmatchNet architecture: enhancing CLRNet with MatchNet integration. Gray blocks represent CLRNet's original components, and the pink block depicts MatchNet, which replaces the classic cost function, indicated as "deleted".}
    \label{fig:CLRmatchNet}
\end{figure*}

\textbf{Row-based Methods} Row-based methods follow a top-to-bottom pipeline, such as \cite{philion2019fastdraw,qin2020ultra, li2019line, LaneATT, CLRNet, CondLaneNet, CLRerNet}, focusing on optimizing the shape of the line through regression of relative coordinates. Predefined anchors mitigate the impact of the no-visual-clue problem, improving instance discrimination. However, fixed anchor shapes limit the degree of freedom to describe line shapes. Yet, these methods are the state-of-the-art approaches. A significant milestone in lane detection was achieved with Line-CNN \cite{li2019line}, which adapted the conventional bounding box to the line shape and introduced a predefined set of line proposals as anchors. In order to predict traffic lines as a whole, they proposed a network that consists of two successive parts - an initial backbone followed by Line Proposal Unit (LPU). Then, each of the predictions was assigned a positive/negative label based on a cost function that consists of the average absolute value of the differences in their common indices of sampled rows and utilizes a fixed threshold to divide the positives and negatives. LaneATT \cite{LaneATT} integrated a lightweight CNN backbone together with an anchor-based attention mechanism and expanded the label assignment cost function to include classification scores. CondLaneNet \cite{CondLaneNet} presents a novel approach of conditional lane detection. This strategy is based on conditional convolution and a row anchor-based formulation. Specifically, it initially identifies the starting points of lane lines before proceeding with row anchor-based lane detection. 
CLRNet \cite{CLRNet} introduced a novel network architecture that leverages both low-level and high-level features and proposed the Line IoU metric to better regress lines. They adopted the SimOTA method, as proposed by YOLOX \cite{YOLOX}, where for each GT lane, the k top predictions with the lowest cost were chosen as positive samples and the other predictions were labeled as negative samples. The detector dynamically selected the maximum number k of positive matches for each GT based on the LineIoU metric \cite{CLRNet}. Additionally, they reformulated the cost function between prediction $i$ and GT $j$ as follows:
\begin{flalign}
\label{eq:costfunction}
    \begin{gathered}
    C_{ij} = \omega_{sim}(C_{dis_{ij}} \cdot C_{\theta_{ij}} \cdot C_{(xy)_{ij}})^2+ \omega_{cls}C_{cls_{ij}}
    \end{gathered}
\end{flalign}

Where $C_{dis}{}$ is the regression cost of all valid lane points between GT $j$ and prediction $i$. $C_{\theta}{}$ is the difference in the mean angle, $C_{xy}{}$ is the regression cost of the starting point coordinates and $C_{cls}{}$ is the focal cost \cite{FocalLoss}. $\omega_{sim}{}=-3$ and $\omega_{cls}=1{}$ are the weight parameters for the geometric similarity term and 
the classification term, respectively, and were set likely as a result of a simple heuristic approach. 

CLRerNet \cite{CLRerNet} introduced a novel definition of IoU called LaneIoU, which takes into account local angles, with the goal of providing a more accurate representation of the IoU metric when calculating the overlap of the segmentation mask during the testing phase. They reformulated the cost function by substituting the geometric attributes originally used in CLRNet with the LaneIoU loss as follows: 
\begin{equation}
    \label{eq:clrer_costfunction}
    C_{ij} = -LaneIoU_{ij}+\lambda C_{cls_{ij}} \tag{2}
\end{equation}
Where $\lambda$ is a weighting parameter.

While adapting the parameters to better define the lane correlation for the cost function is important, an equally crucial aspect is the establishment of relationships between these parameters when defining the cost function. We propose a deep-learning-based label assignment method that employs the same cost parameters (Eq. \ref{eq:costfunction}) utilized by CLRNet and seeks to acquire an optimal label-assignment method.

\section{Method}
\subsection{Motivation}
Modern anchor-based algorithms commonly employ straight lines as anchors, regressing them to obtain desired lane shapes. This approach introduces challenges, particularly for curved lanes, which often have fewer high-quality predictions corresponding to them, whereas straight lanes tend to have more.
The initial geometric alignment between predictions and curved GT lanes tends to be low, potentially leading the label assignment method to favor predictions with high confidence, but poorly aligned with curved GT lanes as positive matches. This bias might result in high classification scores for predictions with low IoU with GT lanes. 

\subsection{CLRmatchNet}
We adopted the state-of-the-art CLRNet as the baseline of our model due to its superior performance compared to alternative methods. To enhance its capabilities, we incorporated MatchNet to replace CLRNet's original classic label assignment method. The architectural framework of the network is presented in Fig. \ref{fig:CLRmatchNet}. CLRNet components are colored gray, while our network, MatchNet, is colored pink. 

The model utilizes ResNet\cite{resnet} or DLA\cite{DLA} as a backbone, generating three levels of feature maps using the feature pyramid network (FPN) \cite{FPN}. Each level of the feature map is then pooled by $N$ lane priors, defined by $(x_{0}, y_{0},\theta)$, where $x_{0}$ and $y_{0}$ indicate the starting point of the line, and $\theta$ represents the line's angle. For the deepest layer $L_{0}$, these priors are uniformly distributed, and for the two following layers ($L_{1}, L_{2}$) they are refined and learned. The refined lane priors and the feature maps are the input to the ROIGather module, where each feature map is pooled by its lane priors set to produce a lane feature maps. Followed by convolutions and fully connected model, ROIGather outputs $N$ lane predictions. A lane prediction consists of learned geometric features, including foreground and background probabilities, the length of the lane prior, the starting point of the lane prior, the angle between the x-axis of the lane prior, and $N$ offsets, that is, the horizontal distance between the prediction and its lane prior. For more details, refer to \cite{CLRNet}. 

As a pre-processing step, we trained MatchNet to determine the optimal score for each prediction-GT pair  and classify them as positive or negative matches based on predefined parameters that describe their correlation. During the training phase, we use the pretrained sub-module of MatchNet to assign the $N$ lane predictions to the GTs. MatchNet assigns a score to each candidate pair, resulting in distinct and well-separated scores. For each GT, the network effectively selects the appropriate number of pairs, allowing for a more precise and efficient lane detection process. This dynamic selection method leverages the discriminating power of MatchNet's scoring system and optimizes the match pairing for enhanced performance. Then, the positive matches are fed into the overall loss function used during training that is composed of several components:

\begin{equation}
    \label{loss}
    \tag{3} \mathcal{L} = \lambda_{0}\mathcal{L}_{xyl\theta} + \lambda_{1}\mathcal{L}_{cls} + \lambda_{2}\mathcal{L}_{seg} + \lambda_{3}\mathcal{L}_{LineIoU}
\end{equation}

Where $\mathcal{L}_{xyl\theta }$ is smooth-L1 loss to regress the anchor parameters $(x_{0}, y_{0}, \theta)$ and the length $l$, $\mathcal{L}_{cls}$ is the focal loss \cite{CULane_F1} considering both positive lanes and negative lanes, $\mathcal{L}_{seg}$ is an auxiliary cross-entropy loss for per-pixel segmentation mask, and $\mathcal{L}_{LineIoU}$ is the LineIoU loss. 

\begin{figure*}[t]
    \begin{center}
    \includegraphics[width=1\linewidth]{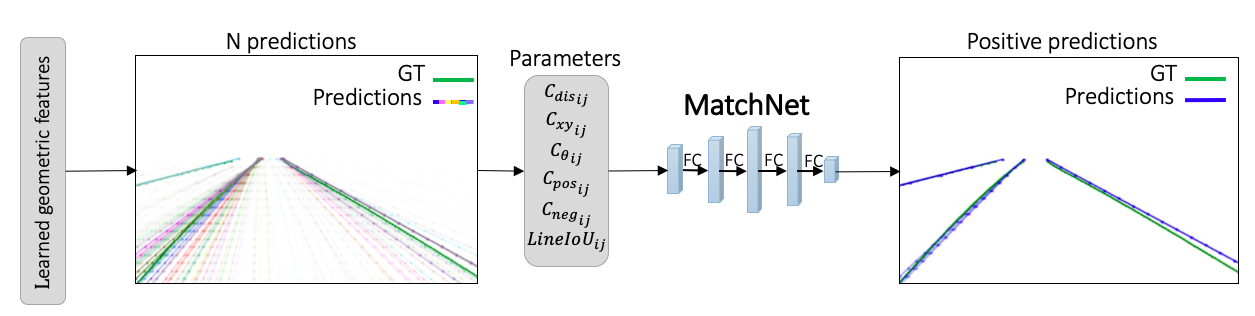}
    \end{center}
    \caption{Architecture of MatchNet. A fully connected model for match classification. As an input, MatchNet receives geometrical parameters of $N$ predictions in relative to each one of the GT, and outputs probabilities defining whether the each pair is positive or negative.}
    \label{fig:matchnet}
\end{figure*}

\subsection{MatchNet}
\subsubsection{Architecture} Fig. \ref{fig:matchnet} presents the architecture of MatchNet. MatchNet is an encoder-decoder model, where each of the stages consists of two FC layers, followed by a leaky ReLU activation function. The last fully connected layer is followed by a sigmoid activation function. MatchNet outputs probabilities for each GT-prediction pair, indicating whether they can be matched or not. The binary cross-entropy loss is adopted for the training process.

\subsubsection{Inputs} MatchNet input consists of the same geometric $(C_{dis_{ij}}, C_{\theta_{ij}}, C_{xy_{ij}})$ and classification parameters $(C_{pos_{ij}}, C_{neg_{ij}})$ used in the classic cost function (Eq. \ref{eq:costfunction}) and the LIoU cost \cite{CLRNet} defined in CLRNet. These parameters are computed for each GT-prediction match candidate. All parameters are normalized to a range between 0 and 1. Additionally, any distances exceeding the size of the image (for x and y distances) or angles greater than 180 degrees (for $\theta$) are clamped to prevent extreme values from negatively affecting the performance of the model.

\subsubsection{Targets} To train MatchNet, we need to label pairs of detected lanes and GT lanes as positive and negative. This process includes two steps. First, we use the classical cost function presented in Eq. \ref{eq:costfunction} to determine whether a match is plausible. Then, we further validate the match by considering the loss produced by CLRNet as valuable feedback for the pair match. In other words, a pair of GT prediction is considered as a match only when two conditions are met: 1) the classical cost function, $C_{ij}$ (Eq. \ref{eq:costfunction}), indicates a positive match between the prediction $i$ and the GT $j$ lane. 2) The loss function calculated by CLRNet $ \mathcal{L}_{ij}$ for that specific pair should be lower than a predefined threshold $t_{L}$. Only when both conditions are satisfied, the match is considered valid and marked as a positive match in the GT data for MatchNet.
\begin{flalign}
    \begin{gathered}
    \label{eq:gtgeneration}
        \tag {4} match_{ij} = 
        \begin{cases} \text{$ +1$}, & \text{if }   (C_{ij}=+1) \cap                                   (\mathcal{L}_{ij}<t_{L})        \\
                    \text{$ -1$}, & \text{else } 
        \end{cases} \\
    \mathcal{L}_{ij} = \mathcal{L}_{LIoU}(ij) + \lambda \mathcal{L}_{cls}(ij)
    \end{gathered}
     \end{flalign}
Where $\mathcal{L}_{LineIoU}$ is the LIoU loss, $\mathcal{L}_{cls}$ is the focal loss and $\lambda=0.1$ is a balancing parameter between them.

\begin{figure*}[t]
    \begin{center}
    \includegraphics[scale=0.35]{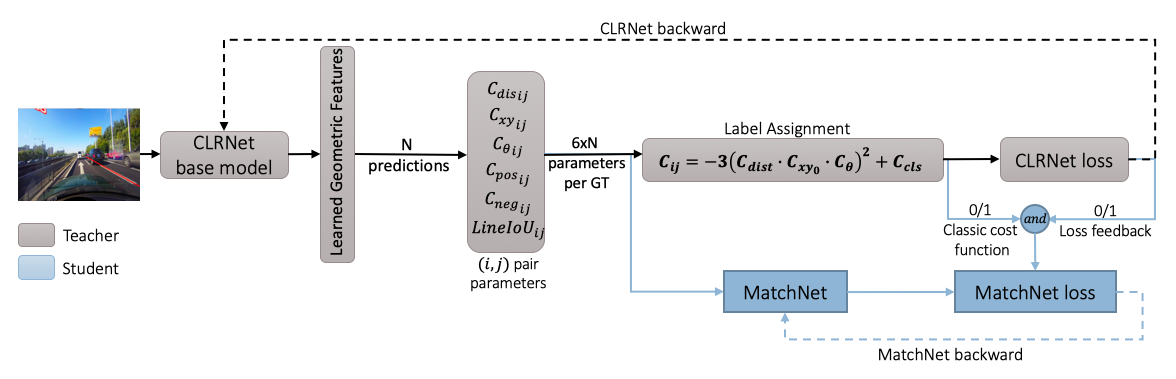}
    \end{center}
    \caption[MatchNet teacher-student training setup]{MatchNet teacher-student training setup. The gray blocks represent the teacher model (CLRNet) and the lower blue blocks represent the student model (MatchNet). In this setup, for each image, the teacher predicts N lanes and computes the cost parameters between each prediction i and each GT j in the image. These parameters are used as input to MatchNet. To determine if the pair (i, j) is a match, we utilize both the classical cost function decision and a feedback from the loss. Only when both criteria are met (set to 1), we designate this pair as a match for reference. MatchNet is trained using the inputs and references supplied by the CLRNet.}
    \label{fig:techearstudent}
\end{figure*}

\subsubsection{Training Method} To train MatchNet, a teacher-student approach is adopted, where CLRNet acts as the teacher model and MatchNet acts as the student model. This configuration is shown in Fig. \ref{fig:techearstudent},  divided into two parallel components: the teacher (CLRNet) shown in gray blocks and, at the bottom, the student (MatchNet) represented in blue blocks. The teacher, CLRNet, is tuned for 3 additional epochs while generating both the input data and the reference target to the student, MatchNet. In this setup, MatchNet learns the decision-making process of the teacher, benefiting from the strengths of the classical cost function while incorporating the knowledge acquired from CLRNet. 

\subsubsection{Dynamic k Selection} CLRNet incorporates the SimOTA \cite{YOLOX} assigner to dynamically select the number of prediction matches for each GT, denoted as $k$. The value of $k$ is determined by summing the LineIoUs of positive anchors and clipping the sum between 1 and $k_{max}=4$. Our approach introduces a dynamic selection mechanism for $k$ based on the scores assigned by MatchNet. Only matches that exceed a predefined confidence threshold are classified as positive matches. 
This innovative strategy ensures that the matching process is not solely dictated by the LineIoU metric, but is also influenced by the quality of each detected lane.

\subsection{Curves Subset}
In our experiments we used the CULane dataset \cite{CULane_F1} that stands as the standard benchmark for lane detection tasks. CLRNet exhibited suboptimal performance in scenarios involving curved lanes, a specific category within the test dataset. Aware of the need for improvement, we devised a strategy to improve lane detection, especially in these challenging conditions. We filtered a specific training dataset consisting of 8,677 images that exclusively feature curved lanes, and we trained both MatchNet and CLRmatchNet on this subset. First, MatchNet was trained using this subdataset, and subsequently, we fine-tuned CLRNet's results using the pre-trained MatchNet on the same subdataset. This approach allowed us to leverage the knowledge learned by MatchNet to further refine CLRNet's performance in the curve section.


\section{Experiments}
\subsection{Datasets}\label{subsec:datasets}
In our experiments, we conducted research using the CULane dataset  \cite{CULane_F1}, a widely used benchmark for lane detection. This dataset contains 100,000 images, divided into three subsets: training set, validation set, and test set. All images are formatted to have dimensions of 1640x590 pixels. The test set is categorized into 9 groups, with 8 of them considered challenging cases such as challenging curved lanes and difficult lighting conditions such as shadows and night frames.
Specifically, we employed the curved subset extracted from the training set of the CULane dataset to train our models (MatchNet and CLRmatchNet).

\subsection{Implementation Details}
\subsubsection{CLRmatchNet} Our implementation details are completely identical to the baseline CLRNet \cite{CLRNet}. The adopted approach utilizes pre-trained ResNet \cite{resnet} or DLA \cite{DLA} as backbone architectures. The input images were resized to dimensions of 320x800. 
The optimization process incorporates the AdamW \cite{Adam} optimizer and a cosine learning rate decay \cite{Decay}.
For training, we initialize CLRmatchNet weights with the pre-trained CLRNet's weights and incorporate the pre-trained MatchNet's weights for an additional 5 epochs to refine the label assignment. We adopted a threshold of 0.7 for MatchNet's output score during training, categorizing pairs with scores that surpass this threshold as positive and those that below it as negative. MatchNet is a tiny neural network and had minimal impact on the overall training time.
For testing, we adapted the initial threshold of CLRNet that filter predictions to be included in the F1 score calculation from 0.40 to 0.43.

\begin{table*}[t]
    \begin{center}
    \resizebox{1\textwidth}{!}{
        \begin{tabular}{cccccccccccccccc}
            \hline 
            Method & Backbone & mF1 & $\text{F1}_{50}$ &  $\text{F1}_{75}$  & \textbf{Curve} & Normal & Crowd & Dazzle & Shadow & Noline & Arrow & Cross &  Night &   GFLOPs & FPS \\
            \hline
            LaneATT \cite{LaneATT} & Res34 & 49.57 & 76.68 & 54.34 & 67.72 &  92.14 & 75.03 & 66.47 & 78.15 & 49.39 & 88.38  & 1330 & 70.72 & 18.0 &  170 \\
            CondLane\cite{CondLaneNet}  & Res34 & 53.11 & 78.74 & 59.39 & 73.88 & 93.38 &  77.14 &  71.17 & 79.93 & 51.85 & 89.89 & 1387 & 73.92 &   19.6    &    237\\
            CLRNet\cite{CLRNet} & Res34 & 55.14 & \textbf{79.73} & \textbf{62.11} & 72.77 & 93.49 & 78.06 & \textbf{74.57} & \textbf{79.92} & 54.01 & \textbf{90.59}  & \textbf{1216} & 75.02 & 21.5 &  204 \\
            \hline
            CLRmatchNet (ours)& Res34 & \textbf{55.22}& 79.60 & 62.10 & \textbf{75.57} &     \textbf{93.49}\textbf{ }& \textbf{78.52} & 74.56 & 79.62 &  \textbf{54.23} &     90.55  & 1898 & \textbf{75.03} &  21.5 & 204 \\
            \hline
             &  &  &  &  &  &  &  &  &  &  &  &  &  &  &
        \end{tabular}}
    \\
    \resizebox{1\textwidth}{!}{
    \begin{tabular}{cccccccccccccccc}
        \hline
        Method & Backbone & mF1 & $\text{F1}_{50}$ &  $\text{F1}_{75}$ & \textbf{Curve} &Normal & Crowd & Dazzle & Shadow & Noline & Arrow & Cross &   Night & GFLOPs & FPS \\
        \hline
        CondLane\cite{CondLaneNet}  & Res101 & 54.83 & 79.48 & 61.23  & 75.21 & 93.47 & 77.44 & 70.93 & 80.91 & 54.13 & 90.16 & \textbf{1201}&   74.80 &   44.8 & 97\\
        CLRNet\cite{CLRNet} & Res101 & 55.55 & \textbf{80.13} & 62.96 &  75.57 & \underline{\textbf{93.85}} & 78.78 & \textbf{72.49} & 82.33 & 54.50 &   89.79 & 1262 & \underline{\textbf{75.51}} & 42.9 & 94 \\
        \hline
        CLRmatchNet (ours)& Res101 & \underline{\textbf{55.69}} & 80.00 & \underline{\textbf{63.07}} & \underline{\textbf{77.87}} & 93.82 & \textbf{79.04} & 71.55 & \textbf{82.38} & \underline{\textbf{55.03}} & \textbf{89.90 } & 1610 & 74.84 &  42.9 & 94 \\
        \hline
         &  &  &  &  &  &  &  &  &  &  &  &  &  &  &\\
    \end{tabular}}
    \\
    \resizebox{1\textwidth}{!}{
    \begin{tabular}{cccccccccccccccc}
        \hline
        Method & Backbone & mF1 & $\text{F1}_{50}$ &  $\text{F1}_{75}$ & \textbf{Curve} & Normal & Crowd & Dazzle & Shadow & Noline & Arrow  & Cross & Night & GFLOPs & FPS \\
        \hline
        CLRNet\cite{CLRNet} & DLA34 & \textbf{55.64} & \underline{\textbf{80.47}} & \textbf{62.78} & 74.13 & 93.73 & \underline{\textbf{79.59}} &       \underline{\textbf{75.30}} & \underline{\textbf{82.51}} & 54.58 & 90.62  & \underline{\textbf{1155}} & \textbf{75.37} &  18.4 & 185 \\
        \hline
        CLRmatchNet (ours)& DLA34 & 55.14 &  79.97 & 62.10 & \textbf{77.09} & \textbf{93.74} & 79.50 & 73.57 & 82.27 & \textbf{54.61} & \underline{\textbf{90.96}} &  2064 & 75.19 &  18.4 & 185\\
        \hline
    \end{tabular}}
    \label{table:culane_results}
    \caption{Evaluation results on the CULane test set. \textbf{Scores} (bold) indicate the best results achieved on the same backbone, while \underline{\textbf{scores}} (bold and underline) represent the overall best results across all backbones.} 
    \end{center}
\end{table*} 

\subsubsection{MatchNet} The selection of inputs plays an important role in ensuring a well-balanced dataset for effective and unbiased training. During the training of MatchNet, CLRNet generates predictions and their associated parameters relative to the GT. We select all matches classified as positive by CLRNet's classical label assignment method from the first image in each batch of teacher training. To achieve a balanced dataset, additional random negative lane matches are chosen to complete the set to a total of batch size of 18 positive and negative examples. We used a threshold of $t_{L}=0.3$ for the loss feedback. 
For training in a teacher-student mechanism, we used the pre-trained weights of CLRNet as a teacher and trained MatchNet on the curve subsets for 5 epochs. The optimization process incorporates the AdamW optimizer with a learning rate of 0.0001 and a cosine learning rate decay.

\subsection{Evaluation Metric} 
The official metric of the CUlane dataset in the evaluation process is the F1 score \cite{CULane_F1}. During evaluation, first predictions with confidence scores exceeding a predefined threshold are filtered. These selected predictions are then utilized to calculate the final F1 score which is derived from the IoU metric. Since the IoU relies on areas instead of points, a lane is represented as a thick line connecting the respective lane's points. In particular, the official metric considers the lanes as 30-pixels-thick lines. If a prediction has an IoU greater than $t_{IoU}=0.5$ with a GT lane, it is considered a TP, whereas unmatched predictions and GTs are referred to as $FP$ and $FN$. The F1 score is then calculated as follows:

\begin{equation}
\label{eq:F1score}
\tag{5} F_1=\frac{2\times Precision \times Recall}{Precision+Recall}
\end{equation}

\section{Results}
\subsection{CLRmatchNet Results}
\begin{figure}[h]
    \begin{center}
    \includegraphics[width=0.9\linewidth]{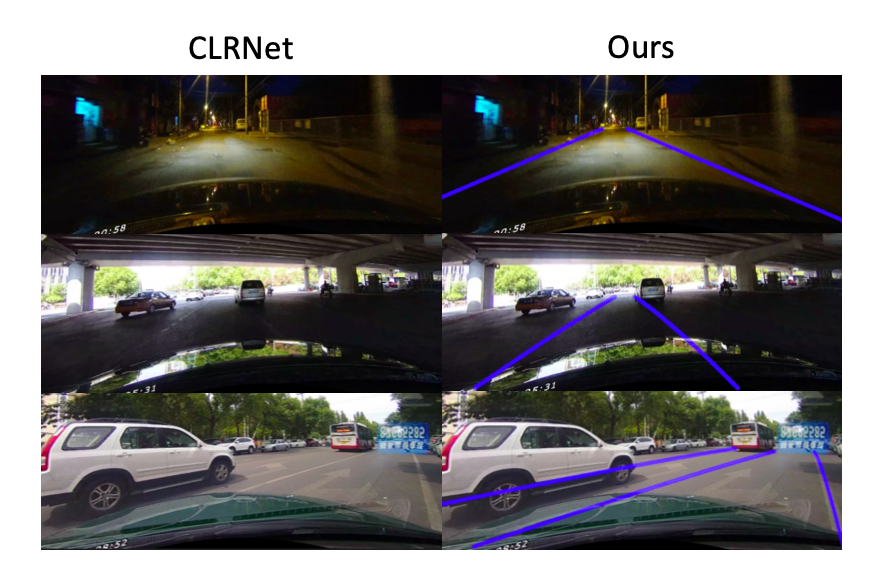}
    \end{center}
    \caption{Visualization results of CLRNet and our method for night, shadow, and crowd scenarios of the CULane dataset (from top to bottom)}
    \label{fig:results}
\end{figure}

We present the results of our approach on the CULane lane detection benchmark dataset and perform a comparative analysis with other well-established lane detection techniques. Tab. \ref{table:culane_results} summarizes our findings, showing that our proposed method (CLRmatchNet) has achieved a significant improvement in the curve section of the CULane dataset, resulting in $F1_{50}$ scores of 75.57\% for ResNet34 (+2.8\%), 77.87\% for ResNet101 (+2.3\%) and 77.09\% for DLA34 (+2.96\%). Furthermore, our approach demonstrates enhanced performance in the 'crowded' section for ResNet34 (+0.46\%) and the 'no-line' section (+0.53\%) for ResNet101, while maintaining comparable results for other sections. We show qualitative results of our method on the CULane dataset compared to CLRNet in Fig. \ref{fig:results} for the following scenarios: night, shadow, and crowd (from top to bottom). 

It is important to note that in this dataset, in our experiments, the results for night, cross and dazzle conditions had an impact on the overall performance. However, we believe that this impact might be due to the quality of the GT in these scenes. Basically, MatchNet should not behave differently in these scenarios, and we assume that these scenes will also be improved. See Sec. \ref{sec:discussion} for further analysis.

\begin{figure}[t]
    \begin{center}
    \includegraphics[width=0.7\linewidth]{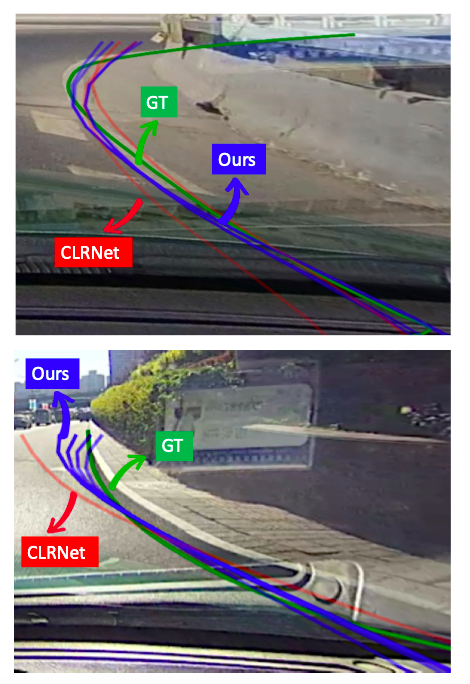}
    \end{center}
    \caption{Comparison of label assignment results: classic CLRNet assignment vs our  approach (MatchNet), better viewed in colors. Color intensity represents confidence. CLRNet's and MatchNet's positive assignments are colored red and blue, respectively. MatchNet outperforms CLRNet in selecting lanes with both better confidence and geometric alignment.}
    \label{fig:assignment}
\end{figure}

\subsection{MatchNet Results} Fig. \ref{fig:assignment} illustrates the results of the assignment process for a curved lane of both CLRNet and our approach, MatchNet. The color intensity of the predictions corresponds to their confidence scores. Lanes identified as positives by CLRNet's label assignment method are shown in red, whereas those identified as positives by MatchNet (our approach) are shown in blue. GT lanes are colored green. 

Noticeably, the prediction assigned as positive by CLRNet's assignment method (red) has both a low-confidence score and a suboptimal geometric alignment. In contrast, our method excels in identifying a superior match for this GT, with a higher-confidence score and better geometric alignment. This example highlights that the conventional label assignment technique initially employed does not consistently produce the optimal matches that we aim for the network to learn from.
 
\subsection{Boosting Confidence}
\begin{figure}[h]
    \begin{center}
    \includegraphics[width=0.8\linewidth]{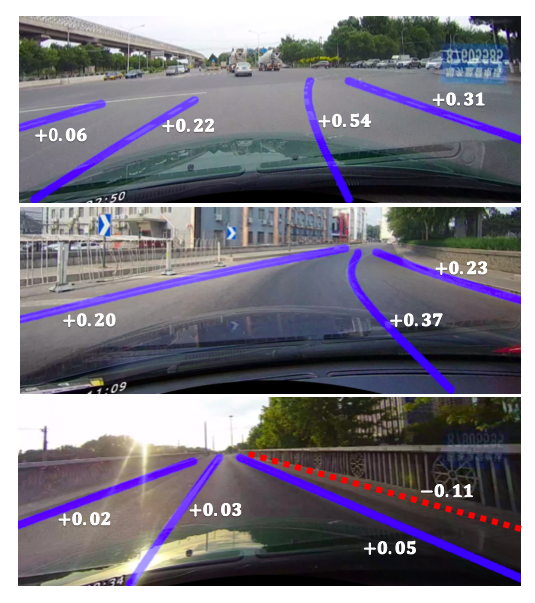}
    \end{center}
    \caption[Enhanced lane detection confidence: comparative analysis of CLRmatchNet and CLRNet]{Enhanced lane detection confidence: comparative analysis of CLRmatchNet and CLRNet. The improvement in confidence scores for each lane detection in CLRmatchNet compared to CLRNet is expressed as $\Delta x$.}
    \label{fig:scores}
\end{figure}

On evaluation, only lane detections that exceed a predefined confidence threshold are classified as valid detections, forming the basis for calculating the F1 score (Eq. \ref{eq:F1score}). Employing MatchNet as the assignment method has yielded a remarkable improvement in the confidence levels of detection. Fig. \ref{fig:scores} illustrates the improvements in detection confidence scores in the range [0,1], denoted as $\Delta x$. This visualization indicates a significant increase in confidence for all detected lanes. Additionally, a red dashed line indicates a decrease in the confidence level of lanes falsely detected by CLRNet. 

\begin{figure}[h]
    \begin{center}
    \includegraphics[width=1\linewidth]{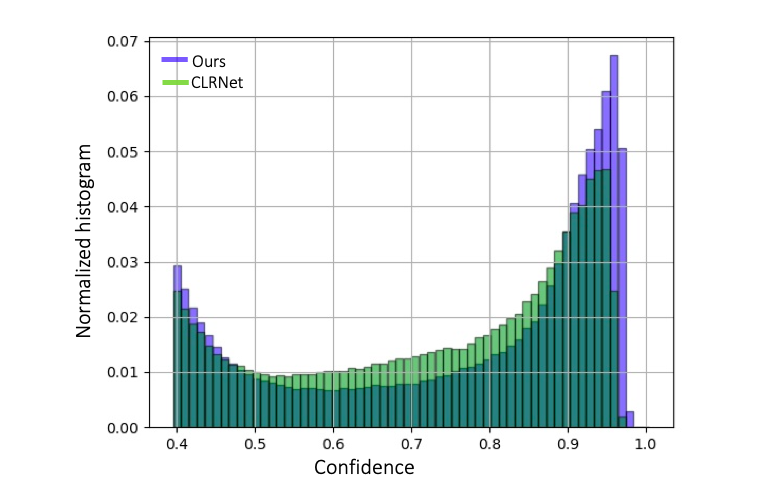}
    \end{center}
    \caption[Normalized histogram comparison of confidence scores: CLRmatchNet vs. CLRNet]{Normalized histogram comparison of confidence scores: CLRmatchNet vs. CLRNet. Our model, CLRmatchNet (purple), shows notably higher confidence scores compared to CLRNet (light green), resulting in a more pronounced distribution of scores. Shared areas are depicted in dark green.}
    \label{fig:histograms}
\end{figure}

Fig. \ref{fig:histograms} presents the normalized confidence scores histograms for both CLRNet (green) and our proposed CLRmatchNet method (purple). Our model shows higher confidence scores in lane detections compared to the original distribution. Additionally, our model's confidence scores follow a more parabolic distribution, in contrast to the original distribution, which was notably more uniform. This change in confidence scores implies that our CLRmatchNet has a stronger grasp on accurate lane detection. This boost in confidence scores has enabled us to elevate the confidence threshold during evaluation, thereby ensuring a more reliable and accurate evaluation of our model's performance.

\section{Ablation Study}
Tab. \ref{table:ablation} presents the results of fine-tuning the CLRNet model weights on ResNet101 for an additional 5 epochs, incorporating both the curve sub-dataset and MatchNet as label assignment method for improving the performance on the curve section. As can be seen, the results in the curve section of CLRNet initially yielded a score of $F1_{50}=75.57\%$. Subsequent fine-tuning on the curve dataset over 5 epochs slightly improves the results and achieves $F1_{50}=75.90\%$. Then, when we refined the model by fine-tuning it with both the curve dataset and a pre-trained MatchNet as the label assignment model, we managed to increase the initial $F1_{50}$ score by 2.3\% to 77.87\%. 

\begin{table}[h]
    \begin{center}
    \begin{tabular}{cccccc}
    \hline
    CLRNet & curved dataset & MatchNet& $\text{F1}_{50}$ &   \\
    \hline
    \checkmark & - & -  & 75.57 \\
    \checkmark & \checkmark & -  & 75.90 \\
    \checkmark & \checkmark  & \checkmark & \textbf{77.87}\\
    \hline
    \end{tabular}
    \end{center}
    \caption{Refining CLRNet on the curve subset using MatchNet for ResNet 101}
    \label{table:ablation}
\end{table}

\section{Discussion} \label{sec:discussion}
\subsubsection{Inaccurate GT Annotations} Although CLRmatchNet significantly improves the results for the curve section, it negatively affects the results of several sections and does not improve overall performance. Our method detects many more TP lanes and lowers the FN rate in all sections compared to the CLRNet results with the initial confidence threshold for the test. However, we are experiencing more FP detections, especially in the cross section. It is noteworthy that the dataset we used might not have been labeled optimally. Our method has shown the ability to identify lanes not originally labeled as GT. 

\begin{figure}[t]
\begin{center}
    \centering
    \includegraphics[width=1\linewidth]{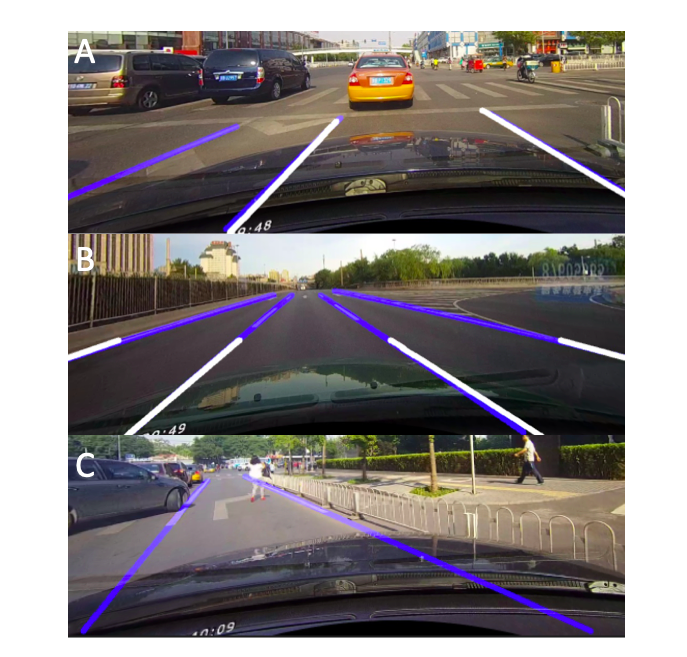}
    \caption[Inaccurate GT annotations]{Inaccurate GT annotations. Visualization of frames in which our model correctly identifies lanes that were not marked as GTs, resulting in false positives. The detections of our model are highlighted in blue and the GT lanes are represented in white.}
    \label{fig:gtlabel}
    \end{center}
\end{figure}

Fig. \ref{fig:gtlabel} illustrates three scenarios in which our model successfully detected true lanes that were not marked as GT, resulting in them being marked as FP. Our model's detections are highlighted in blue, and the GT lanes are highlighted in white. In the upper frame, the left-most lane is clearly a real lane, and our network detects it as one, however it was not marked as a GT. In the middle frame, the GTs are very short, but the actual lanes are much longer. Our network accurately identifies the entire length of the lane. In the lower frame, none of the lanes was marked as GT. In all these cases, the network excels in detecting real lanes; however, due to discrepancies in GT and network predictions, these instances are classified as FP, significantly influencing the final F1 score.

\subsubsection{MatchNet Convergence} During MatchNet training, we observed instances where the convergence exhibited fluctuations in the loss function for certain training examples. However, it is important to note that despite these fluctuations, the overall loss in the training process improved as expected. This may be the result of qualitative labeling of the data, as the labels were not assigned manually, but rather through an automated process. This automated labeling approach might introduce some variability into the training process, leading to occasional fluctuations in the loss. Nevertheless, our training strategy effectively improved CLRNet's performance, resulting in improved lane detection results in the curves section. 

\subsubsection{Comparative Analysis} 
It is worth noting that CLRerNet\cite{CLRerNet} exhibits superior results compared to both CLRNet and our model (CLRmatchNet). Both studies, CLRNet and ours, identify label assignment as a weakness of CLRNet, particularly for curved lanes. A key distinction lies in the methodologies employed: CLRerNet is based on classical methods, while our approach is a deep learning-based approach.

\section{Conclusions}
We presented MatchNet, a novel approach designed to improve the label assignment process in the context of lane detection. Integrated into the state-of-the-art lane detection network CLRNet \cite{CLRNet}, MatchNet replaces the conventional label assignment with a submodule deep neural network, resulting in the CLRmatchNet model. Our contributions are multifaceted: firstly, we introduce a deep learning-based label assignment method, addressing limitations inherent in classical cost functions. In particular, our approach significantly enhances the detection capability of curved lanes while maintaining comparable performance across other testing categories. Moreover, the confidence levels of true positive lanes see a substantial boost, enabling the adjustment of the confidence score threshold used in the evaluation process. Our method introduces dynamic lane matching flexibility, allowing dynamic selection of maximum matches per GT based on MatchNet scores. 

Classical cost functions are well-established, while our innovative label assignment approach highlights the potential for improvements across various object detection algorithms, beyond the scope of lane detection, that depend on classical cost functions.

{\small
\bibliographystyle{ieee_main}
\bibliography{main}
}
\end{document}